\documentclass[conference]{IEEEtran}
\IEEEoverridecommandlockouts
\usepackage{cite}
\usepackage{amsmath,amssymb,amsfonts}
\usepackage{amssymb}
\usepackage{amsfonts}
\usepackage{algorithmic}
\usepackage{graphicx}
\usepackage{textcomp}
\usepackage{xcolor}
\usepackage{multirow}
\def\BibTeX{{\rm B\kern-.05em{\sc i\kern-.025em b}\kern-.08em
    T\kern-.1667em\lower.7ex\hbox{E}\kern-.125emX}}

\author{
    \IEEEauthorblockN{Ruoyu Wang\IEEEauthorrefmark{1}, Wenqian Wang\IEEEauthorrefmark{2}, Jianjun Gao\IEEEauthorrefmark{1}, Dan Lin$^{\dagger{\rm ,}}$\IEEEauthorrefmark{3}, Kim-Hui Yap\IEEEauthorrefmark{1}, Bingbing Li\IEEEauthorrefmark{4}}
    \IEEEauthorblockA{\IEEEauthorrefmark{1}School of Electrical and Electronic Engineering, Nanyang Technological University, Singapore}
    \IEEEauthorblockA{\IEEEauthorrefmark{2}Continental-NTU Corporate Lab, Nanyang Technological University, Singapore}
    \IEEEauthorblockA{\IEEEauthorrefmark{3}College of Computer Science and Technology, Harbin Engineering University, China}
    \IEEEauthorblockA{\IEEEauthorrefmark{4}Continental Automotive Singapore Pte. Ltd., Singapore}
}

\begin{document}

\title{MultiFuser: Multimodal Fusion Transformer for Enhanced Driver Action Recognition\\

}

\maketitle

\begin{abstract}

Driver action recognition, aiming to accurately identify drivers' behaviours, is crucial for enhancing driver-vehicle interactions and ensuring driving safety.
Unlike general action recognition, drivers' environments are often challenging, being gloomy and dark, and with the development of sensors, various cameras such as IR and depth cameras have emerged for analyzing drivers' behaviors. 
Therefore, in this paper, we propose a novel multimodal fusion transformer, named MultiFuser, which identifies cross-modal interrelations and interactions among multimodal car cabin videos and adaptively integrates different modalities for improved representations.
Specifically, MultiFuser comprises layers of Bi-decomposed Modules to model spatiotemporal features, with a modality synthesizer for multimodal features integration. Each Bi-decomposed Module includes a Modal Expertise ViT block for extracting modality-specific features and a Patch-wise Adaptive Fusion block for efficient cross-modal fusion. 
Extensive experiments are conducted on Drive\&Act dataset and the results demonstrate the efficacy of our proposed approach.



\end{abstract}


\section{Introduction}

\begin{figure}[tbp]
\centering
\includegraphics[width=0.5\textwidth]{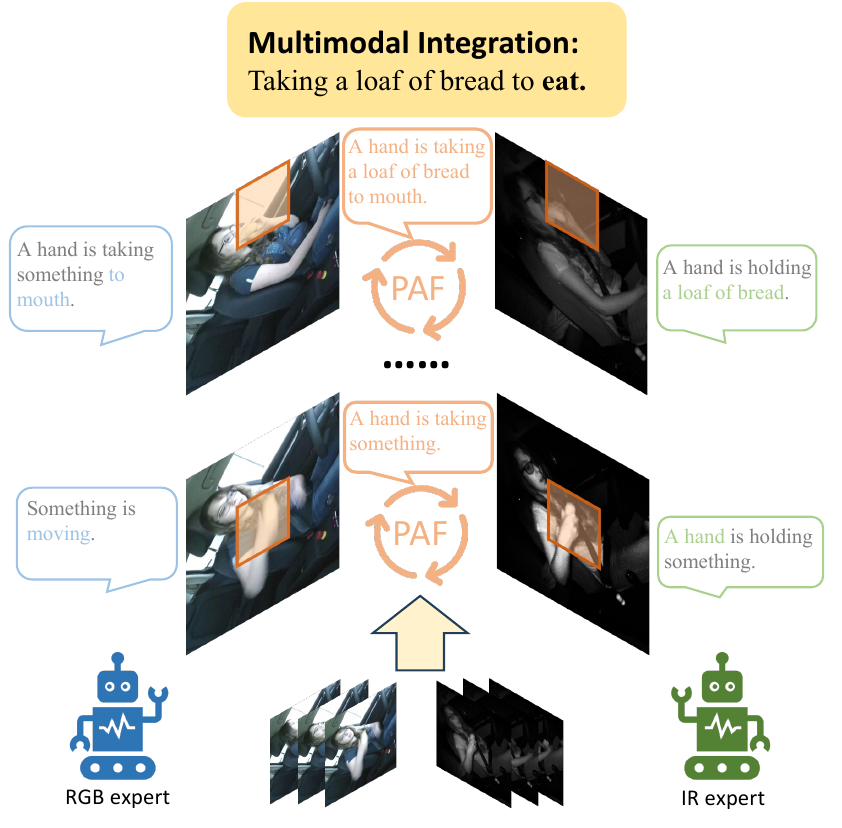}
\caption{
For multimodal video sequences (such as RGB and IR), the Patch-wise Adaptive Fusion (PAF) can integrate information from different modalities on a per-patch basis to form a multimodal feature representation for each patch. 
Subsequently, through Multimodal Integration, all these features are aggregated to achieve a comprehensive understanding of the entire video, which is then utilized for driver action recognition.}
\label{motivation}
\end{figure}

In recent years, Driver Action Recognition (DAR) has received increasing attention.
The target of this task is to automatically identify the driver's action with the videos from the visual sensors in the car cabin environment. 
DAR is essential for advanced driving assistant systems as it enables identifying distractions, recognizing drivers' behavior and predicting their intended actions.
With the advances in autonomous driving technology granting drivers increased freedom of activity, DAR is playing an essential role in enhancing both human-vehicle communication and driving safety.


Benefiting from the rapid development of deep learning architectures~\cite{resnet, Liu1,  gao1}, significant progress has been achieved in the DAR task~\cite{cabinsurvey, transdarc, mifi}. 
Previous methods mostly focus on adopting unimodal videos as input, employing video-based vision backbone models such as Swin Transformer~\cite{swin} to predict drivers' actions.
However, compared to other action recognition tasks, relying solely on single-modality input may be insufficient for DAR.
In most scenarios, RGB is the preferred choice due to its ability to provide a rich, natural color representation, while  InfraRed (IR) and Depth videos mainly capture thermal information and spatial relationships, respectively. 
Nevertheless, under suboptimal lighting conditions, such as those encountered when driving through tunnels or during nighttime, RGB videos may become excessively dark, compromising feature reliability.
In these situations, combining features from IR and Depth videos, which are less sensitive to lighting variations, can significantly enhance the model's accuracy and robustness in action recognition.
Consequently, features extracted from different modalities contain distinct information and emphasize specific attributes, rendering them appropriate for varied scenarios. 
By processing multiple modalities, a model can effectively integrate and leverage these diverse informational elements, thereby enhancing its expressive capabilities and delivering a more nuanced and comprehensive representation of driver actions.
However, the integration of multimodal videos in driver action recognition presents challenges, requiring an effective fusion strategy to develop a more comprehensive representation.

Existing studies~\cite{khan2022supervised, mod4, real, wang1} typically extract spatiotemporal information separately from each modality and integrate them using decision-level fusion techniques. These methods often involve pooling prediction scores from each unimodal stream or concatenating final-layer features across different streams to form a unified output. However, decision-level fusion introduces higher computational complexity and may inadequately capture interactions between modalities.
Vision Transformers (ViTs) have excelled in visual tasks like image classification and object detection. Yet, their application to video tasks is hindered by the added temporal dimension, which significantly increases computational demands and challenges in accurately representing spatiotemporal features, especially with multimodal inputs. 
Moreover, simplistic fusion techniques overlook the inherent interdependencies and complementary nature of information across modalities. Consequently, these approaches fail to capture complex intermodal interactions, leading to less robust and holistic feature representations. Addressing these challenges requires developing fusion strategies that can effectively handle multimodal data and intricate temporal dynamics while enhancing cross-modal interactions.


In this study, we aim to identify the cross-modal interrelations and interactions among multimodal car cabin videos and, based on these insights, adaptively fuse the modalities to achieve a more comprehensive and holistic representation of driver actions.
Thus, we propose a multimodal driver action recognition network that can adaptively extract the modality-specific spatiotemporal features and model their patch-wise interactions into a detailed cross-modal representation.
Specifically, we design layers of a Bi-decomposed Module to learn the unimodal and cross-modal features of driver actions with a modality synthesizer for multimodal features integration.
Firstly, Bi-decomposed Module utilizes a Modal Expertise ViT to leverage the pre-trained backbone in extracting modality-specific features on inter-modality decomposed tokens.
Meanwhile, we propose a Patch-wise Adaptive Fusion (PAF) block in the intra-modality decomposition stream for efficiently capturing the interrelation and interactions among the patches of different modalities, as shown in Fig~\ref{motivation}.
Subsequently, the patch-wise cross-modal features are effectively integrated into a modality synthesizer by a Multimodal Integration block.
Finally, the multimodal feature is concatenated with modality-specific class tokens to form a holistic and comprehensive representation of the DAR classification.


We summarize our contributions as follows.
(1) We propose a novel multimodal driver action recognition model that employs layers of Bi-decomposed Module to effectively extract modality-specific features and patch-wise cross-modal features with a modality synthesizer to integrate them for a comprehensive multimodal feature.
(2) We devise a Patch-wise Adaptive Fusion block to capture the the interrelations among different modalities for detailed cross-modal features and introduce a Modal Expertise ViT that leverages large-scale multimodal pretraining to adaptively extract modality-specific spatiotemporal features.
(3) MultiFuser outperforms the state-of-the-art methods on Drive\&act dataset, showing the effectiveness of integrating features from multimodal video inputs.

\section{Related Work}

Driver action recognition, which aims to identify behaviors of a driver in the car cabin, is different from generalized action recognition task~\cite{dynamicAR, maginet, wang1, li2023token, wang2024todo}.
This task is challenging due to the unstable lighting conditions in the car cabin.
After Martin et al.~\cite{driveact} proposed the large-scale multi-modal driver action recognition dataset Drive\&Act, Peng et al.~\cite{transdarc} introduced the swin-transformer~\cite{swin} into this task and developed a feature calibration system to enhance the generalizability of the model.
More recently, Kuang et al.~\cite{mifi} proposed to integrate multi-view features to reduce the impact of view obstruction.
However, they all focused on only using uni-modality as input in driver action recognition, neglecting the advantages of incorporating multi-modal data.

Benefiting from the development of multimodal models~\cite{cai1, shao1, wang2, cai2}, researchers have also investigated how multimodal video inputs can be used to predict driver actions.
Beyond uni-modal methods, Lin et al.~\cite{dar} proposed an efficient multi-modal model DFS and achieved good performance.
Rotiberg et al.~\cite{mod4} conducted experiments to comparatively evaluate the performance of different decision-level fusion methods across multi-modality inputs.
However, these methods rely on late fusion after extracting features from each modality separately, without paying attention to the interactions and complementary information between different modalities.
Our MultiFuser approach, while preserving the unique characteristics of each modality, emphasizes the connections between features across modalities, resulting in a more comprehensive and clearer multi-modal representation of driver actions.

\section{Methodology}

\begin{figure*}[t]
\centering
\includegraphics[width=1.0\textwidth]{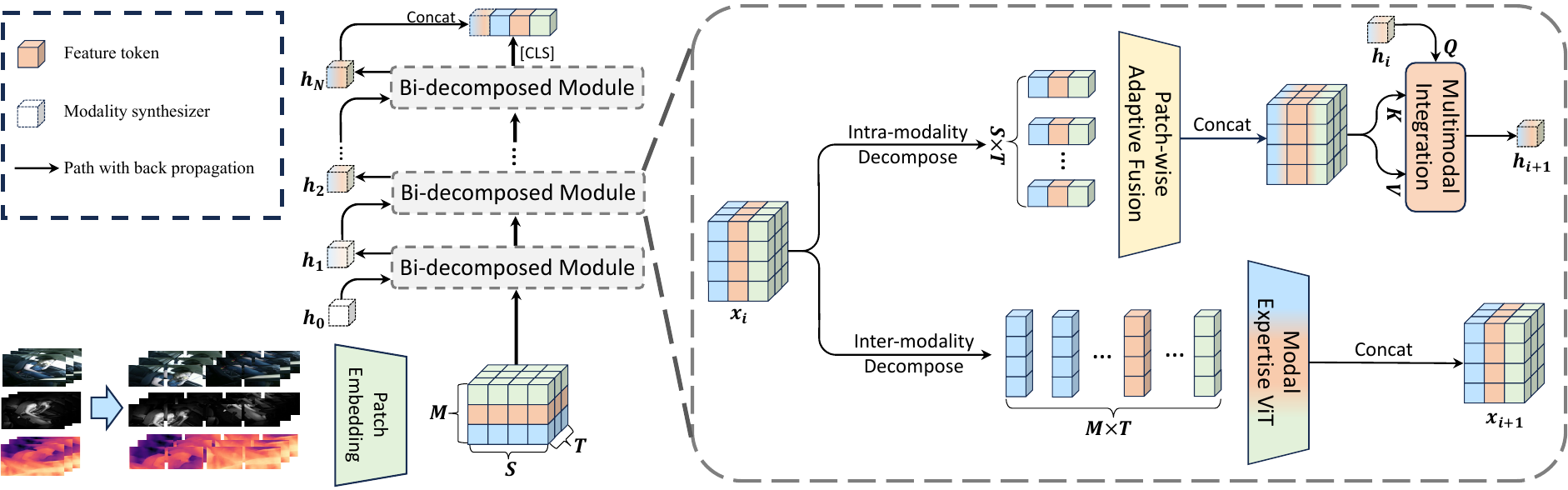}
\vspace{-10pt} \caption{
\textbf{Overview} of MultiFuser, a network proposed to achieve a comprehensive multimodal representation for driver action recognition. 
For multimodal patch tokens input, it designs $N$ layers of a Bi-decomposed Module to model its spatiotemporal features with a modality synthesizer for multimodal features integration.
Each Bi-decomposed Module comprises a Modal Expertise ViT block within inter-modality decomposition to extract the modality-specific features and a Patch-wise Adaptive Fusion block within the intra-modality decomposition stream for efficient cross-modal fusion.
Finally, the modality synthesizer integrates these cross-modal features and then combines with the \texttt{[CLS]} token from each modality, providing a comprehensive and holistic multimodal representation of driver actions.
}
\label{overview}
\end{figure*}

In this section, we present our Multimodal Fusion Transformer (\textbf{MultiFuser}) designed for driver action recognition.
As shown in Fig.~\ref{overview}, MultiFuser adeptly learns and integrates both cross-modal and intra-modal spatiotemporal features from multimodal inputs, yielding a comprehensive and holistic representation. 
We have developed a series of Bi-decomposed Modules, each comprising inter- and intra-modality decomposition streams, which efficiently extract intra-modal features and facilitate their integration into interactive cross-modal features. 
To establish a uniform multimodal representation across different modalities, we introduce a modality synthesizer token that integrates these fused cross-modal features into a multimodal feature. 
Subsequently, the concatenation of class tokens from each modality with the modality synthesizer token provides a detailed depiction of driver actions, offering distinct and valuable insights for subsequent classification.

\subsection{Overview}

Initially, MultiFuser segments the multi-modality video inputs into discrete patches and then linearly embeds them as $ \mathbf{X}_0 \in \mathbb{R}^{M \times S \times T \times D} $, incorporating positional embeddings.
Here, $M$, $S$, $T$ and $D$ represent the modal, spatial, temporal and embedding dimensions respectively,
To efficiently and effectively learn the representation of the multimodal driver action videos, we construct a stack of $ N $ layers of bi-decomposed modules to extract both unimodal and multimodal features of $ X_0 $.
Each bi-decomposed module processes the input $X_i$ through two parallel streams: inter-modality decomposition and intra-modality decomposition. 
In the inter-modality decomposition, we have developed a Modal Expertise ViT specifically designed to efficiently extract modality-specific features. 
Concurrently, we introduce a Patch-wise Adaptive Fusion (\textbf{PAF}) block, which facilitates the fusion of patch-wise multimodal features, thus providing a more detailed and comprehensive depiction of driver actions. 
Beyond the Bi-decomposed Module, we have engineered a novel \textbf{modality synthesizer} to integrate these multimodal features into a unified representation that captures the holistic information pertinent to multimodal inputs. 
Specifically, the modality synthesizer consolidates the fused features from the PAF into a multimodal class token, which distinctly characterizes driver actions for classification purposes. 
Ultimately, the modality synthesizer is concatenated with the class tokens from each modality, and this amalgamated output is subsequently linearly projected to predict the specific actions performed by the driver in the input video.

\subsection{Intra-modality Decompose}

To learn the multimodal representation of driver actions, we develop an intra-modality decomposition stream that is capable of effectively fusing the modality-specific features and integrating the cross-modal features.
Initially, the intra-modality decomposition segregates the input $x_i$, retaining solely the modal dimension and discarding the spatial and temporal dimensions:
\begin{equation}
    [\mathbf{B}_i^1, \mathbf{B}_i^2, \cdots, \mathbf{B}_i^{S \times T}] = {\rm IntraDecompose}(\mathbf{X}_i),
\end{equation}
where $\rm IntraDecompose$ refers to the operation of intra-modality decomposition and $B_i^j$ refers to the $j$th intra-modal decomposed embedding of $B_i$.
In other words, each embedding $B^j_i$ consists of $M$ patch tokens from the same spatial position and temporal frame in different modalities.
Thus, by investigating the interrelations between different modality tokens, we can facilitate patch-wise fusion and construct a detailed representation with multimodal patch features.

\subsubsection{Patch-wise Adaptive Fusion}

To achieve a detailed multimodal depiction of driver action videos, we propose the PAF that performs multimodal fusion at a fine-grained, patch-wise level.
Performing patch-wise fusion allows for more focused modeling of the current local information, minimizing interference from inputs at other locations and enabling each patch to obtain a more precise multimodal representation. 
Additionally, by reducing the length of the input sequence, this approach significantly conserves computational resources compared to processing all tokens simultaneously.
For the multimodal patch feature token input, denoted as $\mathbf{B}_i^j$, we initially compute the similarity between each patch to delineate their interrelations.
Subsequently, these interrelations can be adopted as the weights to adaptively fuse the unimodal features for an effective fusion of cross-modal features.

The PAF block keeps the residual structure with layer norm to enhance stability and ensure efficient training.
\begin{equation}
    {\rm PAF}(\mathbf{B_i^j}) = \Phi({\rm Norm} (\mathbf{B_i^j}))+\mathbf{B_i^j}.
\end{equation}
As illustrated in Fig.~3, the adaptive fusion is based on the interrelation between the different patch feature tokens to weigh their values and we also follow the multi-head strategy of classical Transformer for comprehensive representation.
\begin{equation}
    A_n(\mathbf{B}) = R_n V_n(\mathbf{B}),
\end{equation}
\begin{equation}
    \Phi(\mathbf{B}) = {\rm Concat}(A_1(\mathbf{B});A_2(\mathbf{B}); \cdots; A_n(\mathbf{B})U,
\end{equation}
where $A_n(\cdot)$ refers to the weighted features in the n-th head.
$R_n$ is an affinity matrix that describes token interrelation and $V_n(\cdot)$ is a linear projection, while $U \in \mathbb{R}^{C \times C}$ is a learnable fusion matrix.
As for the interrelation calculation, we calculate the dot product between the target token and each patch token as a measure of similarity and then normalize these similarity values to determine the extent of interrelation with each patch token.
For the interrelation value between the target token $\mathbf{B}_m$ and the token $\mathbf{B}_t$, it is computed as follows:
\begin{equation}
\mathrm{R}_n\left(\mathbf{B}_m, \mathbf{B}_t\right)=\frac{\exp \left\{\mathrm{Q}_n\left(\mathbf{B}_m\right)^T \mathrm{~K}_n\left(\mathbf{B}_t\right)\right\}}{\sum_{t^{\prime}=1}^M \exp \left\{\mathrm{Q}_n\left(\mathbf{B}_m\right)^T \mathrm{~K}_n\left(\mathbf{B}_{t^{\prime}}\right)\right\}},
\end{equation}
where $Q_n(\cdot)$ and $K_n(\cdot)$ are different linear projections of the $n$th head.
With these interrelation values, the model can adaptively fuse the unimodal patch features into a unified cross-modal patch representation.
Finally, these outputs are activated by a feed-forward network and then concatenated to form $X'_i$, which serves as the input for further integration.


\subsubsection{Multimodal Integration}

Unimodal features are effectively fused within the PAF to construct a multimodal representation for each patch. 
Subsequently, we designed the Multimodal Integration Block to reweight and integrate the cross-modal features of all patches into the modality synthesizer token. 
This process enables the unique token to comprehensively summarize the classification information for the entire multimodal video.

Considering the temporal, spatial, and modal variability of multimodal videos, we have implemented dynamic positional embeddings prior to integration. 
Specifically, we employ depth-wise convolution for positional encoding, as it accommodates inputs of varying lengths and is more conducive to multimodal videos while preserving the absolute positional information of each token.
\begin{equation}
    \hat{\mathbf{X}}_i  = {\rm DWConv(\mathbf{X}'_i)}+\mathbf{X}'_i,
\end{equation}
where $\rm DWConv$ means simple 3D depthwise convolution with zero paddings.
After the activation from the dynamic positional embedding, the modality synthesizer is proposed to integrate the multiple patch-wise cross-modal features into a unique representation of the driver action information.
To be more specific, we follow the multi-head cross attention to compute the attention between the modality synthesizer $\mathbf{h}_i$ and the cross-modal features $\hat{\mathbf{X}}_i$, and then use the attention to multiply the linearly projecting values of $\hat{\mathbf{X}}_i$ to achieve information integration.
\begin{equation}
    \mathbf{h'}_i = {\rm MHCA}({\rm Norm}(\mathbf{h}_i),{\rm Norm}(\hat{\mathbf{X}}_i))+\mathbf{h}_i,
\end{equation}
\begin{equation}
    \mathbf{h}_{i+1} = {\rm FFN}({\rm Norm} (\mathbf{h'}_i))+\mathbf{h}_{i+1},
\end{equation}
where $\rm Norm$ refers to the layer norm and $\rm FFN$ refers to the linear feed-forward network.
Following this process, the patch-wise cross-modal features are integrated into the modality synthesizer, endowing it with the capability to provide a comprehensive representation of all multimodal input videos. 
This enhanced representation supplies richer multimodal features in the subsequent prediction phase, aiding the model in accurately determining the driver's actions.

\subsection{Inter-modality Decompose}

To efficiently extract the spatiotemporal features of each modality input, we design the inter-modality decomposition stream.
Similarly, we first apply an inter-modality decomposition to isolate the spatial dimension by removing the modal and temporal dimensions.
The input $\mathbf{X}_i$ is decomposed into a series of embeddings $ \mathbf{F}_i^k \in \mathbb{R}^{S\times D} $.
\begin{equation}
    [\mathbf{F}_i^1, \mathbf{F}_i^2, \cdots, \mathbf{F}_i^{M \times T}] = {\rm InterDecompose}(\mathbf{X}_i),
\end{equation}
where $\rm InterDecompose$ denotes the operation of inter-modality decomposition.
Through this decomposition, each embedding in the series corresponds to a complete frame representation, meaning that all tokens within a single embedding are derived from the same video frame.
This approach offers two significant benefits: (i) substantially reducing the length of the input sequence, mitigating the quadratic complexity that later stages of the Vision Transformer (ViT) would otherwise struggle to manage, and (ii) preserving the spatial relationships between tokens in each input, leveraging the strong generalization capabilities of the pre-trained ViT backbone.

\subsubsection{Modal Expertise ViT}

Given the distinct feature information and relationships inherent in different modalities, we have designed the Modal Expertise ViT to adaptively learn the representations of each modal input.
The structure of Modal Expertise ViT is a vanilla ViT $\mathcal{V} (\cdot)$ with a group of different parameter sets.
Firstly, we utilize the same CLIP ViT checkpoint to initialize all of 
the parameter sets for the powerful capacity of CLIP~\cite{clipvit}.
Subsequently, we fine-tune the Modal Expertise ViT on the multimodal classification dataset NTU RGB+D~\cite{nturgbd}. 
This process enables each distinct set of parameters to generalize the learning of spatial relationships specific to each modality and to effectively extract the corresponding features.
Thus, Modal Expertise ViT is able to extract the modality-specific unimodal features of each video input with suitable parameters.
For the patch tokens input $\mathbf{F}_i^k$, the Modal Expertise ViT chooses the corresponding $p$th set of parameters to learn the features of it.
\begin{equation}
    \mathbf{F}_{i+1}^k = \mathcal{V}_p(\mathbf{F}_i^k),
\end{equation}
\begin{equation}
    \mathbf{X}_{i+1} = {\rm Concat}_{k=1}^{M \times T}(\mathbf{F}_{i+1}^k).
\end{equation}
The outputs are subsequently concatenated to form $X_{i+1}$, which serves as the input for the next layer of the Bi-decomposed Module.

\section{Experiments and Results}
\subsection{Implementation Details}


For each video input, we sample $8$ frames and resize each frame to $224\times224$.
These frames are then split into $8\times16\times16$ patches for embedding.
Our network architecture consists of $N=12$ layers of the Bi-decomposed module, and we follow the baseline UniFormerV2~\cite{uniformerv2} that the modality synthesizer is employed in the last $K=4$ layers.
The modality synthesizer and the $M$ class tokens of single-modality are concatenated and fed into a fully connected layer followed by a softmax layer for classification.
We employ AdamW~\cite{adamw} as the optimizer in $100$ epochs training with an initial learning rate of $1\times10^{-4}$.
Before fine-tuning on DAR dataset, the Modal Expertise ViT is first pretrained on the NTU RGB+D dataset~\cite{nturgbd} for $100$ epochs.

\subsection{Dataset and Metrics}

The Drive\&Act~\cite{driveact} is a large-scale (over 9.6 million frames) video dataset in DAR. 
This dataset encompasses eight types of multi-modal input data: RGB, IR, Depth and Near-InfraRed (NIR) from five different perspectives
It categorizes activities into three distinct levels: scenarios, fine-grained activities, and atomic action units.
For the scope of this paper, our experiments are focused on classifying RGB, IR, and Depth videos taken from the right-top view into their respective fine-grained activities.
We adhere to the three predefined splits from the dataset and average the outcomes to obtain the final result.
To evaluate our results, we utilize two metrics: Mean-1 Accuracy (average per-class accuracy) as the primary metric, and Top-1 Accuracy for implementation assessment.

\subsection{Experimental Results}

To assess the performance of MultiFuser, we conducted a series of experiments on the Drive\&Act dataset. 
Firstly, we benchmarked MultiFuser against prevailing single-modality and multi-modality state-of-the-art models. 
Specifically, to underscore the significance of multi-modal inputs on accuracy, we tested the MultiFuser model using different input modalities. 
Furthermore, we contrasted various fusion strategies to validate the efficacy of our fusion technique. 
Lastly, we also investigated the impact of different model structures on the experimental outcomes.

\subsubsection{Overall results of MultiFuser}

\begin{table}[t]
\caption{The overall results of MultiFuser in comparison with existing single-modality and multi-modality methods (Accuracy in \%).}
\begin{center}
\resizebox{0.48\textwidth}{!}{
\begin{tabular}{llcc}
\hline
Methods & Modality & Mean-1 & Top-1 \\ 
\hline
C3D~\cite{c3d} & NIR & - & 43.41 \\
P3D~\cite{p3d} & NIR & - & 45.32 \\
CTA-NET~\cite{ctanet} & NIR & - & 65.25 \\
I3D~\cite{i3d} & IR  & - & 64.98 \\
TSM~\cite{tsm} & IR & 59.81 & 67.75 \\
TransDARC~\cite{transdarc} & RGB &  60.10 & 76.17 \\
UniFormerV2~\cite{uniformerv2} & RGB & 61.79 & 76.71  \\
\hline
ResNet~\cite{resnet} & IR, Depth & 51.08 & 56.43     \\ 
TSM~\cite{tsm} & IR, Depth & 61.11 & 70.31 \\ 
UniFormerV2~\cite{uniformerv2} &  RGB, IR, Depth & 61.58 & 78.63 \\
MDBU (I3D)~\cite{mod4} & IR, NIR & 62.02 & 76.91 \\ 
DFS ~\cite{dar} & IR, Depth & 63.12 & 77.61 \\
\textbf{MultiFuser (Ours)} & RGB, IR, Depth & \textbf{70.67} & \textbf{82.39} \\ 
\hline
\end{tabular}
\label{SOTA table}
}
\end{center}
\end{table}

We first compare our MultiFuser with existing models on the Drive\&Act dataset (results in Table.~\ref{SOTA table}). 
Since these methods are primarily single-modal, we supplement some models~\cite{resnet, tsm, uniformerv2} using the late fusion strategy on RGB, IR, and Depth modalities. 
According to the table, MultiFuser significantly outperforms all methods, achieving a Mean-1 accuracy of 70.62\% and a Top-1 accuracy of 82.39\%. 
This Mean-1 accuracy surpasses the SOTA single-modality CTA-NET by 8.9\% and exceeds the multi-modality MDBU by 8.65\%.
Moreover, the Top-1 accuracy outperforms the SOTA single-modality by 5.68\% and the multi-modality by 3.76\%.

\subsubsection{Results on different modalities}

\begin{table}[t]
\caption{Ablation study the different modalities on MultiFuser.}
\begin{center}
\resizebox{0.48\textwidth}{!}{
\begin{tabular}{l l c c}
\hline
Fusion Methods & Modality & Mean-1 & Top-1 \\ 
\hline
\multirow{7}{*}{\textbf{MultiFuser}} 
&RGB & 61.79 & 76.71 \\
&IR & 59.64 & 72.56 \\
&Depth & 56.35 & 69.21 \\
&RGB+IR & 69.04 & 80.68 \\
&RGB+Depth & 66.43 & 79.90 \\
&IR+Depth & 64.42 & 76.65 \\
&RGB+IR+Depth & \textbf{70.67} & \textbf{82.39} \\
\hline
\end{tabular}
\label{Modality table}
}
\end{center}
\end{table}

We evaluate MultiFuser results on various combinations of RGB, IR and Depth (results in Table.~\ref{Modality table}).
We observe that in our model, an increase in the input modalities can enhance the recognition accuracy. 
Consequently, when three modalities are input, the model’s accuracy peaks, exhibiting a 10.34\% (Mean-1 accuracy) enhancement compared to the single-modality RGB. 
This demonstrates the effectiveness of our model in performing multimodal fusion and also suggests that the model has the potential to perform better when processing additional modal inputs.
Another observation is the higher model accuracy when RGB is included as an input. 
We postulate this could be attributed to two factors: the rich features provided by RGB data, which are crucial for DAR tasks, and the fact that the ViT structure in the model, having been pre-trained on RGB data, performs optimally in extracting RGB features. 
The overall results underscore the efficacy of utilizing multi-modal data for DAR tasks.

\subsubsection{Results of different fusion strategies}

\begin{table}[t]
\caption{Ablation study the different fusion strategies.}\vspace{-10pt}
\begin{center}
\resizebox{0.35\textwidth}{!}{
\begin{tabular}{l l c c}
\hline
Fusion Methods& Mean-1 & Top-1 \\ 
\hline
Early fusion &  49.88 & 67.30 \\
Late fusion & 61.58 & 78.63 \\
\textbf{MultiFuser cascade} & 62.28 & 79.32 \\
\textbf{MultiFuser parallel} &\textbf{70.67} & \textbf{82.39} \\
\hline
\end{tabular}
\label{fusion methods table}
}
\end{center}
\end{table}

\begin{figure}[t]
\centering
\includegraphics[width=0.48\textwidth]{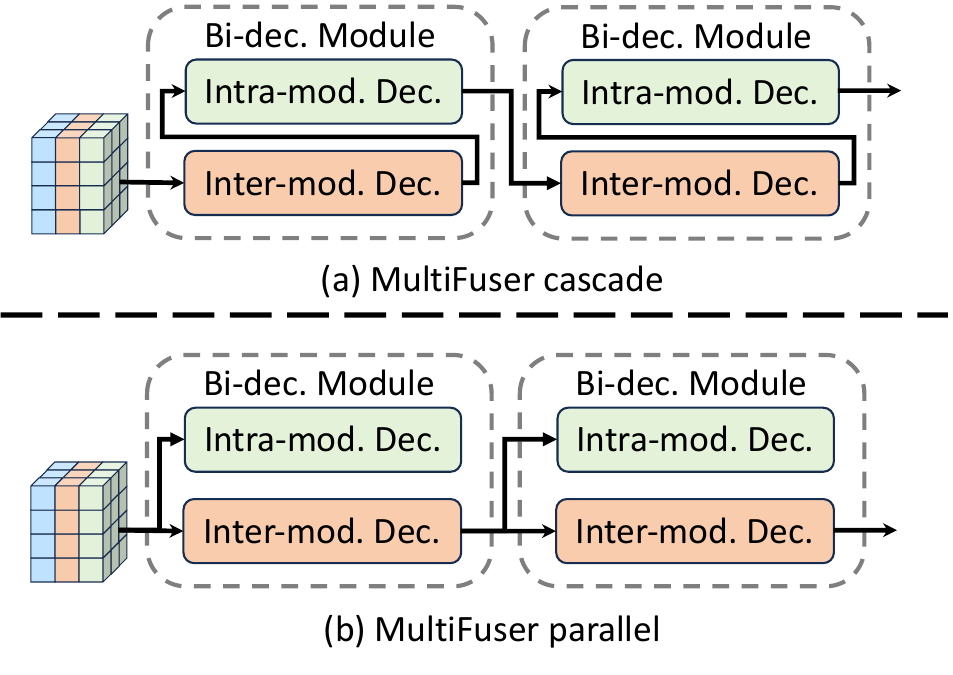}
\vspace{-10pt}\caption{Different connection structure in Bi-decomposed Module. 
(a) MultiFuser cascade extracts the unimodal features (in Inter-modality Decomposition) first and then fuses them (in Intra-modality Decomposition). The fused features are then input into the next Bi-decomposed Module for further unimodal features extraction and fusion.
(b) MultiFuser parallel designs a parallel structure to extract the unimodal and cross-modal features simultaneously which is able to keep the modality-specific features untouched.
}\vspace{-15pt}
\label{fblocation}
\end{figure}

To explore effective fusion strategies while using the same backbone, we conduct a comparative analysis involving four distinct fusion strategies: early fusion, late fusion, MultiFuser cascade and MultiFuser parallel (results in Table.~\ref{fusion methods table}. 
In the early fusion approach, multiple modal data are concatenated at the embedding stage, proceeding as a unified input into the model. 
In contrast, the late fusion utilizes multiple streams to separately extract different modality features, which are then unified in a fully connected layer to derive the final classification features. 
We further examine the impact of different connection structures in the Bi-decomposed Module with the same blocks, as illustrated in Fig.~\ref{fblocation}. 
Two configurations were tested: (i) a cascade configuration where unimodal and multimodal feature extraction blocks are sequenced, with features first extracted via inter-modality decomposition and then processed through intra-modality decomposition to extract multimodal features (MultiFuser cascade); (ii) a parallel configuration where unimodal and multimodal feature extraction blocks operate concurrently, ensuring that the unimodal feature extraction stream remains undisturbed by multimodal fusion (MultiFuser parallel).

As shown in Table~\ref{fblocation}, both configurations of our MultiFuser model outperform the results of early and late fusion methods, thereby validating the efficacy and advancement of our fusion approach. 
Furthermore, the MultiFuser results with the parallel configuration significantly surpass those of the cascade configuration. 
We hypothesize that this improvement may be attributed to the parallel configuration’s preservation of the independence of the unimodal feature extraction stream. 
If modal fusion were to occur between layers of unimodal feature extraction, it could disrupt the intrinsic information of the unimodal data, leading to less rich extraction of unimodal spatiotemporal features and negatively impacting the final predictive performance.
Consequently, in our MultiFuser, we employ a parallel structure for the bi-decomposed module.

\section{Conclusion}

In this work, we have developed a novel multimodal DAR model MultiFuser, which effectively learns the modality-specific features and adaptively models the multimodal features for improved representations.
The MultiFuser designed the Bi-decomposed Module to decompose the feature tokens in two different dimensions, facilitating efficient learning.
Specifically, the Modal Expertise ViT in the inter-modality decomposition stream leveraged the well-pretrained backbone to extract modality-specific features. 
Meanwhile, the Patch-wise Adaptive Fusion in intra-modality decomposition captured the interrelations among modalities and employed them for adaptive fusion of cross-modal features.
Furthermore, MultiFuser introduced a novel modality synthesizer to integrate the patch-wise cross-modal features into a holistic multimodal feature and then combine it with the modality-specific class tokens to form a heterogeneous and comprehensive representation for DAR.
MultiFuser achieves stat-of-the-art performance on the Drive\&Act dataset, demonstrating its superior and robust performance in the DAR task.


\section*{Acknowledgement}
This study is supported under the RIE2020 Industry Alignment Fund - Industry Collaboration Projects (IAF-ICP) Funding Initiative, as well as cash and in-kind contribution from the industry partner(s).

{
\bibliographystyle{ieeetr}
\bibliography{ref}
}

\end{document}